\documentclass[lettersize,journal]{IEEEtran}
\usepackage{amsmath,amsfonts}
\usepackage{algorithmic}
\usepackage{algorithm}
\usepackage{array}
\usepackage[caption=false,font=normalsize,labelfont=sf,textfont=sf]{subfig}
\usepackage{textcomp}
\usepackage{stfloats}
\usepackage{url}
\usepackage{verbatim}
\usepackage{graphicx}
\usepackage{cite}
\usepackage[colorlinks=true, linkcolor=blue, citecolor=blue]{hyperref}
\DeclareGraphicsExtensions{.png, .pdf}
\begin{document}

\title{MORPH Wheel: A Passive Variable-Radius Wheel Embedding Mechanical Behavior Logic for Input-Responsive Transformation}

\author{JaeHyung Jang, JuYeong Seo, Dae-Young Lee, Jee-Hwan Ryu
\thanks{JaeHyung Jang, JuYeong Seo, and Jee-Hwan Ryu are with the Department of Civil and Environmental Engineering, Korea Advanced Institute of Science and Technology, Daejeon 34141, South Korea. (e-mail: jhjang.kd@kaist.ac.kr; iamjoong9@kaist.ac.kr; jhryu@kaist.ac.kr).}

\thanks{Dae-Young Lee is with the Department of Aerospace Engineering, Korea Advanced Institute of Science and Technology, Daejeon 34141, South Korea. (e-mail: ae$\_$dylee@kaist.ac.kr).}}

\maketitle




\begin{abstract}
This paper introduces the Mechacnially prOgrammed Radius-adjustable PHysical (MORPH) wheel, a fully passive variable-radius wheel that embeds mechanical behavior logic for torque-responsive transformation. Unlike conventional variable transmission systems relying on actuators, sensors, and active control, the MORPH wheel achieves passive adaptation solely through its geometry and compliant structure. The design integrates a torque-response coupler and spring-loaded connecting struts to mechanically adjust the wheel radius between 80 mm and 45 mm in response to input torque, without any electrical components. The MORPH wheel provides three unique capabilities rarely achieved simultaneously in previous passive designs: (1) bidirectional operation with unlimited rotation through a symmetric coupler; (2) high torque capacity exceeding 10 N with rigid power transmission in drive mode; and (3) precise and repeatable transmission ratio control governed by deterministic kinematics. A comprehensive analytical model was developed to describe the wheel's mechanical behavior logic, establishing threshold conditions for mode switching between direct drive and radius transformation. Experimental validation confirmed that the measured torque-radius and force-displacement characteristics closely follow theoretical predictions across wheel weights of 1.8-2.8kg. Robot-level demonstrations on varying loads (0-25kg), slopes, and unstructured terrains further verified that the MORPH wheel passively adjusts its radius to provide optimal transmission ratio. The MORPH wheel exemplifies a mechanically programmed structure, embedding intelligent, context-dependent behavior directly into its physical design. This approach offers a new paradigm for passive variable transmission and mechanical intelligence in robotic mobility systems operating in unpredictable or control-limited environments.

\end{abstract}


\section{Introduction}
\IEEEPARstart{R}{obotic} systems operating in dynamic environments encounter diverse operational conditions that impose conflicting requirements on their transmission mechanisms: high speed under light loads \cite{2.wensing2017proprioceptive} and high torque under heavy loads \cite{3.garcia2020compact, 4.girard2017leveraging}. Conventional fixed-ratio transmission systems inherently suffer from a trade-off between these two extremes, as they cannot adapt their transmission ratios to changing demands. The limitation is further exacerbated by the narrow efficiency and power ranges of electric actuators, which confine optimal performance to a small torque-speed window. As a result, a significant portion of actuator capacity remains underutilized across varying tasks, leading to overall energy inefficiency and performance degradation. Therefore, there is a critical need for transmission mechanisms that can automatically adapt their torque-speed characteristics to external loading conditions.


\begin{figure}[!t]
    \centerline{\includegraphics[width=7.3cm]{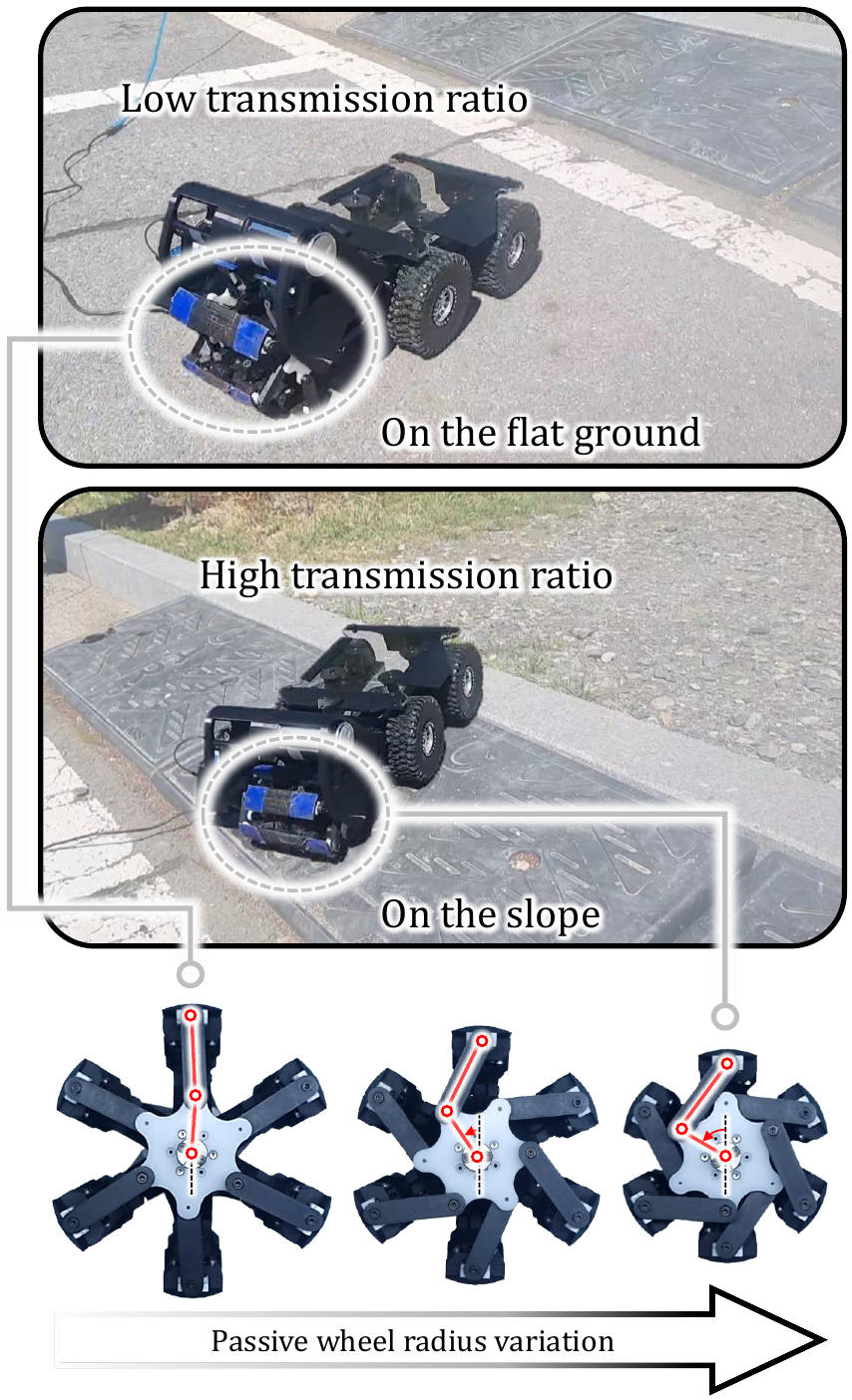}}
    \caption{Mechanically prOgrammed Radius-adjustable PHysical (the MORPH) wheel: Passive variable-radius wheel embedding mechanical behavior logic for input-responsive transformation.}
    \label{fig:fig_1.1}
\end{figure}

\IEEEpubidadjcol Variable transmission (VT) systems offer a promising approach to overcoming the inherent trade-offs of fixed transmission ratio drivetrain by adaptively modulating the torque-speed relationship according to changing load conditions \cite{1.park2024variable}. However, most VT mechanisms implemented in robotic systems rely on active control architectures incorporating sensor, actuators, and feedback algorithms  \cite{jang2022active, 5.chung2024robotic, shin2025variable}. These systems increase overall complexity, consume additional power, and reduce robustness, particularly in field or extreme environments where electronic components are vulnerable. This limitation has motivated interest in passive VT concepts that rely solely on structural responses without electrical actuation. While several passive VT mechanisms have been explored, they have predominantly been applied to tendon-, pulley-, or joint-based actuation rather than wheel-based locomotion. Moreover, existing passive designs for wheels have not yet satisfied critical requirements such as continuous bidirectional operation, high torque transmission capability, and deterministic transformation behavior. These gaps highlight the need for a passive, mechanically programmable wheel capable of robust, fast-response, and torque-responsive transmission behavior.


Recent studies have explored a variety of passive continuously variable transmission (CVT) mechanisms based on deformation-induced structural change, compliant interface transformation, and load-sensitive linkage kinematics. Belter and Dollar proposed a roller-based a passive CVT that alters transmission ratio through variable-pitch contact \cite{7.belter2014passively}, while O'Brien et al. developed an elastomeric transmission that modulates radius by exploiting polyurethane deformation under tensile load \cite{8.o2018elastomeric}. Kim et al. further introduced an elastomer-integrated CVT combined with twisted string actuation \cite{9.kim2020elastomeric}, and Nishinmura et al. designed a passive, compliant linkage-based CVT for high-force robotic grippers \cite{10.nishimura2023lightweight}. Additional works include spring-guided tendon paths for prosthetics \cite{11.chang2024posture}, constant-power ball-disc CVTs \cite{12.cretu2005constant}, and cam-based load-sensitive variable pulleys \cite{13.park2018passively}.


\begin{figure*}[t]
    \centerline{\includegraphics[width=17cm]{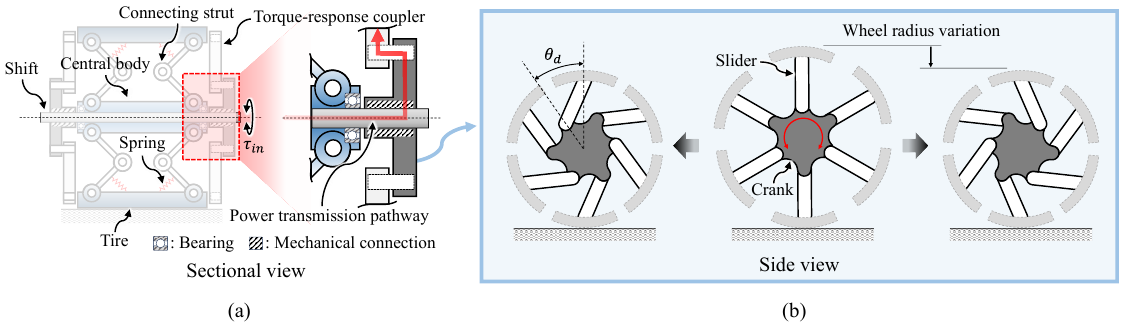}}
    \caption{Mechanically programmed structure of the MORPH wheel. (a) Structural composition of the the MORPH wheel, consisting of a torque-response coupler, connecting struts, springs, tire segments, and central hub. (b) Operating principle of torque-response coupler mechanism.}
    \label{fig:fig_2.1}
\end{figure*}

Although these studies demonstrate meaningful progress, they have been primarily applied to tendon- or joint-driven actuation rather than wheel-driven locomotion, where requirements differ significantly due to continuous bidirectional rotation, high-torque transmission, and accurate transmission ratio design. These constraints are particularly significant given the widespread deployment of wheeled robotic platforms.

Wheeled robots represent one of the most widely used and commercially successful robot platforms due to their maneuverability, energy efficiency, simple control, and suitability for various tasks in both indoor and outdoor environments \cite{17.yu2020quasi, 26.yoon20242, 25.yoon2024stiffness, 27.zheng2022mathbf, 28.kim2023automatic}. Their prevalence is particularly pronounced in infrastructure-limited environments such as space exploration \cite{azkarate2022design}, deep-sea operations \cite{tang2025design}, and underground mining \cite{han2025design}. In these resource-constrained domains, energy efficiency becomes a mission-critical parameter that directly dictates operational success. 

However, conventional wheels with fixed transmission ratios inherently struggle to maintain efficient operation under varying terrain slopes, external loads, and traction conditions, often resulting in excessive motor torque demand or unnecessary energy consumption. A passive CVT is therefore highly desirable for wheeled locomotion, as it passively adjusts the wheel’s effective mechanical advantage in response to terrain-dependent torque loads without requiring additional sensors, actuators, or control effort. 

The origami-inspired variable-radius wheel introduced by Felton et al. is one of the few passive CVT wheel implementations \cite{14.felton2014passive}; however, its unidirectional operation, low torque capacity, and geometry-dependent compliance limit practical deployment in wheeled robotic platforms that operate robustly under diverse terrain, weight, and mission profiles. These constraints indicate that existing passive VT mechanisms do not yet fulfill the above mentioned essential requirements for mobile robotic wheels.

\begin{figure*}[t]
    \centerline{\includegraphics[width=15cm]{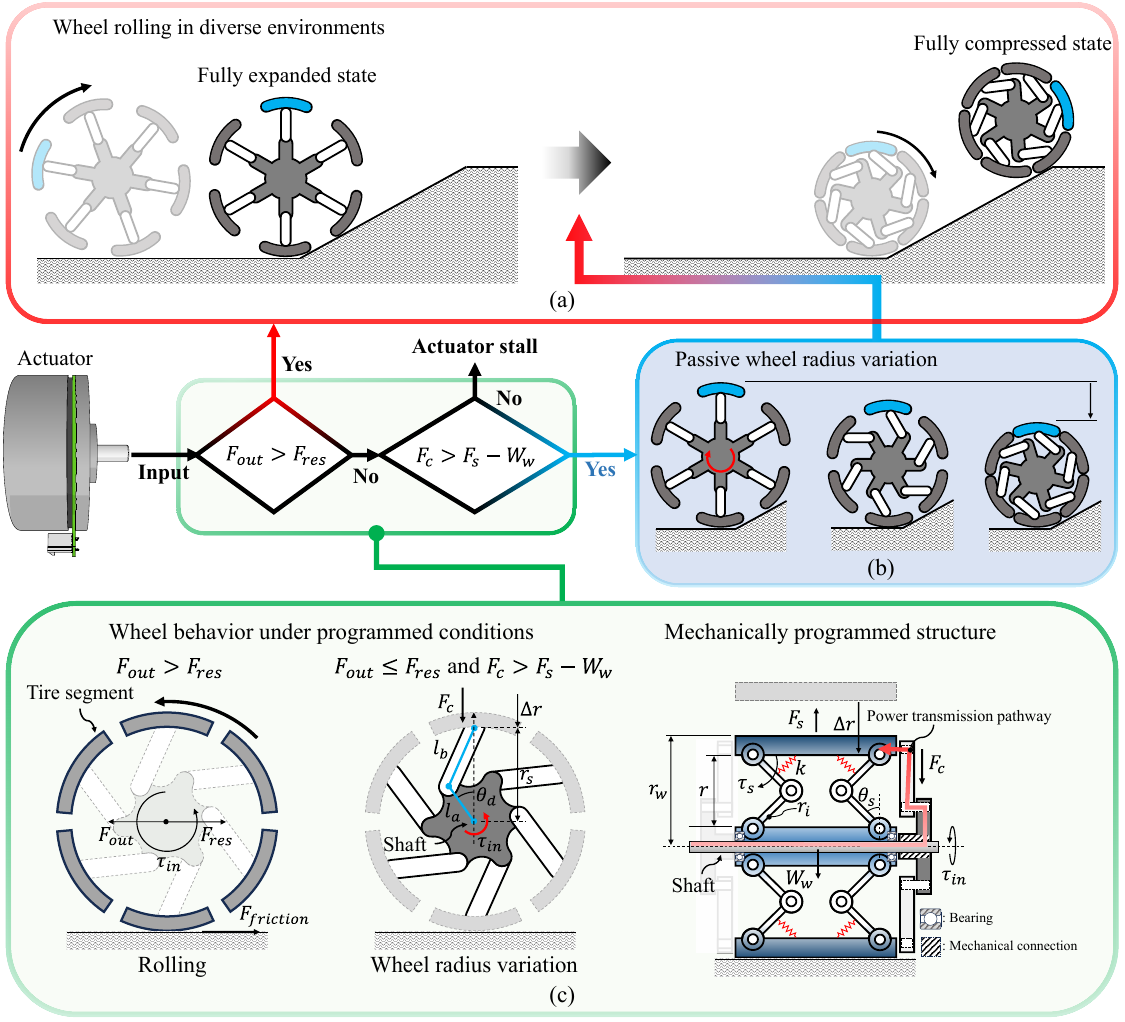}}
    \caption{Mechanical behavior logic of the MORPH wheel:
(a) Wheel rolling in diverse environments: output force is sufficient for direct wheel rotation ($F_{out} > F_{res}$).
(b) Passive wheel radius variation: torque-response coupler reconfigures wheel geometry ($F_{out} < F_{res}$ and $F_c > F_s - W_w$). (c) Wheel behavior under programmed conditions.}
    \label{fig:fig_2.2}
\end{figure*}

To address these shortcomings, we propose the Mechanically prOgrammed Radius-adjustable PHysical (MORPH) wheel-a fully passive continuously variable transmission for wheeled robots (Fig.~\ref{fig:fig_1.1}). At the core of the MORPH wheel design is mechanical programming: decision-making logic is embedded directly into the geometry and compliance of the structure, enabling the wheel to sense input torque and mechanically adjust its radius without sensors, actuators, or electronic control. This mechanically programmed nature is the key enabler of three capabilities that have been difficult to combine in previous designs: (1) bidirectional operation with unlimited rotation, achieved through a symmetric torque-response coupler that works identically in forward and reverse; (2) high torque capacity, maintained by a rigid load path in drive mode so large loads can be transmitted without excessive deformation; and (3) precise and repeatable transmission ratio adjustment, governed by the deterministic kinematics of the coupler-strut-spring mechanism. We validate these capabilities through analytical modeling, bench-top measurements, and mobile robot experiments under varying loads, slopes, and unstructured terrain. Results show that the MORPH wheel expands the effective operating range and improves efficiency of the robotic platform by mechanically adapting its radius to optimize transmission ratio.

The remainder of this paper is organized as follows. Section II introduces the concept of mechanically programmed structures and presents the design principles of the proposed the MORPH wheel. Section III presents the detailed mechanical design of the MORPH wheel, including the configuration of the torque-response coupler, connecting struts, spring-integrated joints, and structural layout optimized for passive radius transformation. Section IV provides experimental validation of the mechanical behavior logic, including measurements of spring resistance and wheel radius variation in response to input torque. Section V demonstrates the performance of the proposed wheel through real-world testing on a mobile robot platform under varying load and terrain conditions. Finally, Section VI concludes the paper with a summary of findings and discussions on future research directions.


\section{Mechanically Programmed Structure}
This paper introduces the concept of a mechanically programmed structure (Fig.~\ref{fig:fig_2.1}), a new design paradigm in which intelligence is embedded directly into the mechanical architecture so that it can adaptively respond to input conditions without any external sensors, actuators, or control systems. In this research, additional degrees of freedom are deliberately incorporated into an underactuated system, and compliant elements are configured to encode behavioral logic through their mechanical properties. As a result, a single input can yield multiple, contextually appropriate outputs based on predefined mechanical thresholds. In this work, we apply this concept for the first time to a wheel, enabling the MORPH wheel to achieve bidirectional operation through symmetric torque sensing, high torque capacity by preserving a rigid load path during normal driving, and precise, repeatable radius adjustment through deterministic coupler-spring kinematics---all without added complexity or loss of robustness.

\subsection{Functional Components of Mechanically Programmed Structure}

The mechanically programmed structure is composed of a set of functional components that collectively enable the system to exhibit programmed mechanical responses without the need for active control. Specifically, the system comprises five primary components as shown in Fig.~\ref{fig:fig_2.1}(a): a torque-response coupler, connecting struts, springs, tire segments, and a central body. Each component is designed to fulfill a distinct mechanical role while their combined operation gives rise to the desired mechanically programmed behaviors. This chapter details the configuration and functionality of these components, highlighting their individual contributions and their interdependent roles within the overall structure.

The torque-response coupler functions as the primary intelligence element of the MORPH wheel, operating as a one-degree-of-freedom mechanism that receives input torque ($\tau_{in}$) from the drive shaft. In contrast to conventional active variable-radius wheels, where input power is transmitted directly to the tire  \cite{20.lee2017origami, 21.lee2021high, 22.moger2024design}, the proposed design uses the coupler to conditionally regulate power transmission according to the magnitude of the input torque. Implemented as a slider-crank mechanism (Fig.~\ref{fig:fig_2.1}(b)), the coupler remains directly connected to the input shaft yet supports two distinct operating modes: (1) conventional torque transfer for wheel rotation and (2) radius variation through geometric reconfiguration.

The connecting struts provide the kinematic linkage between the central body and the tire segments, constraining undesired motion while enabling radial displacement of the tire segments. Each strut integrates a spring element that supplies the compliance required for conditional radius adjustment; these springs are tuned to activate only when the force transmitted from the torque-response coupler exceeds predefined thresholds.

The cental body serves as the structural backbone of the wheel, mechanically supporting the connecting struts and tire segments while remaining isolated from direct torque transmission via bearings. This decoupling ensures that the mechanically programmed behavior of the torque-response coupler functions independently of structural loading, preserving the integrity of the programmed response.

\begin{figure*}[t]
    \centerline{\includegraphics[width=16cm]{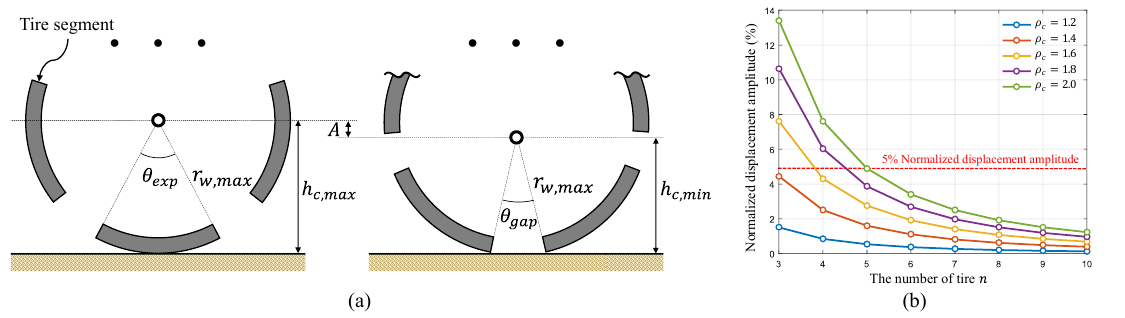}}
    \caption{Wheel displacement analysis for tire segment design. (a) Geometric relationship between tire segments showing the variation in central height ($h_c$) and wheel displacement amplitude $A$ in the fully expanded state. (b) Wheel displacement amplitude as a function of the number of tire segments $n$ and radius ratio ($\rho_c$ = 1.2–2.0), demonstrating that six segments maintain vibration below 5$\%$ of maximum wheel radius.}
    \label{fig:fig_2.3}
\end{figure*}

This configuration embeds mechanical behavior logic: a decision-making framework 
physically encoded within the MORPH wheel's structure, where geometry, compliant 
elements, and kinematics collectively govern the wheel's response to varying 
terrain conditions. Unlike conventional systems that rely on sensors, actuators, or electronic 
control, this logic enables the wheel to respond to input torque and mechanically 
select its operational mode: either maintaining rotation or reconfiguring its 
radius, based solely on its mechanical design. The process operates on a threshold-based principle, where the limits are determined by the stiffness of the compliant elements and the kinematic configuration of the linkages. The fundamental logic structure employs a binary decision tree implemented through mechanical switching mechanisms with two sequential conditions as shown in Fig.~\ref{fig:fig_2.2}. The torque-response coupler functions as the primary logic gate, evaluating input torque $\tau_{in}$ based variables against predetermined mechanical thresholds.


The mechanical behavior logic proceeds through two sequential conditions:\\
\textbf{Condition 1 - Direct rotation mode}: The system first determines whether the output force $F_{out}$ generated by the input torque $\tau_{in}$ exceeds the external resistance force $F_{res}$. If so, the torque-response coupler maintains a rigid kinematic connection between the input shaft and the tire assembly, transmitting all input power into wheel rotation without radius variation.\\
\textbf{Condition 2 - Radius variation mode}: If the first condition is not met, the system evaluates whether the coupler-generated force $F_c$ exceeds the spring resistance $F_s$ minus the wheel's gravitational load $W_w$. When this condition is satisfied, the connecting struts are driven radially inward, reducing the effective wheel radius to increase contraction Force $F_c$. If neither condition is satisfied, the actuator enters a stall state. This hierarchical evaluation ensures stable operation, preventing unintended traditions and enabling the wheel to adapt reliably across diverse load and terrain conditions without active control.




\subsection{Mathematical Modeling of Mechanical Behavior Logic}

The mechanical behavior logic of the MORPH wheel is characterized by a mathematical model based on force equilibrium equations, which specify the threshold conditions for each operational mode. Theses equations formally related the input torque to the force distributions generated within the torque-response coupler and connecting struts. This modeling framework facilitates accurate prediction of mode-switching behavior and supports the optimization of mechanical behavior logic parameters during the design process. In the following, we derive these equations to quantify the relationships between input torque, structural deformation, and the resulting operational state transitions.

In the MORPH wheel, the output force $F_{out}$ is defined by the input torque $\tau_{in}$ and the radius variation $\Delta r$ as follows:
\begin{equation} 
    F_{out} = \frac{\tau_{in}}{r_w}=\frac{\tau_{in}}{(r_i-\Delta r)}
    \label{eq:2.1}
\end{equation}
where $r_w$ is wheel radius and $r_i$ is initial wheel radius. This equation indicates that, unlike a fixed-radius wheel where the relationship between output and input is linear, the MORPH wheel exhibits exponential growth in $F_{out}$ as the input increases. Therefore, calculating $F_{out}$ for a given $\tau_{in}$ requires $\Delta r$ to be expressed as a function of $\tau_{in}$. However, the coupling between input torque $\tau_{in}$ and wheel radius variation $\Delta r$ renders direct calculation infeasible. To establish a tractable mathematical framework for defining output force $F_{out}$, the analysis assumes a given displacement difference $\theta_d$ between the tire segment and input shaft.

Since the MORPH wheel radius variation is governed by the force equilibrium of forces acting perpendicular to the MORPH wheel, $F_c$ is defined by the spring resistance force $F_s$ and the wheel weight $W_w$ as follows:
\begin{equation} 
    F_c=F_s-W_w
    \label{eq:2.2}
\end{equation}
where $F_c$ is determined by the kinematics and dynamics of the torque-response coupler, as shown in the Fig.~\ref{fig:fig_2.2}(c). The geometric relationships of the torque-response coupler are defined as follows:
\begin{equation} 
    \Delta r=(l_a+l_b)-(l_a \cos({\theta_d})+
    \sqrt{{l_{b}}^2-(l_a\sin({\theta_{d})})^2})
    \label{eq:2.3}
\end{equation}
\vspace{0mm}
\begin{equation} 
    \frac{{d\Delta r}}{dt}=\frac{-l_a\sin({\theta_d})(1-(l_a+l_b)\cos({\theta_d}))}{\sqrt{{l_{b}}^2-(l_a\sin({\theta_{d})})^2}}\frac{d\theta_d}{dt}
    \label{eq:2.4}
\end{equation}
where $l_a$ and $l_b$ represent the lengths of the crank and slider in the slider-crank mechanism, respectively. Based on the kinematic relationships, the Jacobian of the torque-response coupler ($J_{coupler}$) and the force required for wheel radius transformation ($F_c$) are defined as:
\begin{equation} 
    J_{coupler}=\frac{-l_a\sin({\theta_d})(1-(l_a+l_b)\cos({\theta_d}))}{\sqrt{{l_{b}}^2-(l_a\sin({\theta_{d}}))^2}}
    \label{eq:2.5}
\end{equation}
\vspace{0mm}
\begin{equation} 
    \tau_{in}=J_{coupler}F_c
    \label{eq:2.6}
\end{equation}
Similarly, the spring resistance force $F_s$ is determined by the kinematics and dynamics of the connecting strut, as depicted in Fig.~\ref{fig:fig_2.2}(c). The geometric relationships of connecting strut are expressed as:
\begin{equation} 
    \theta_s=\arccos{\left( \frac{2r_i-\Delta r}{2r_i} \right)}
    \label{eq:2.7}
\end{equation}
\vspace{0mm}
\begin{equation}
    \Delta r=2r_i(1-\cos({\theta_s}))
    \label{eq:2.8}
\end{equation}
\vspace{0mm}
\begin{equation} 
    \frac{{d\Delta r}}{dt}=2r_i\sin{(\theta_s)}\frac{d\theta_s}{dt}
    \label{eq:2.9}
\end{equation}
where $\theta_s$ and $r_s$ represent link angular displacement and link length of the connecting strut.
The Jacobian of the connecting strut $J_{strut}$ and the spring resistance force $F_s$ are defined as:
\begin{equation} 
    J_{strut}=2r_s\sin({\theta_s})
    \label{eq:2.10}
\end{equation}
\vspace{0mm}
\begin{equation} 
    F_s={J_{strut}}^{-1}\tau_s
    \label{eq:2.11}
\end{equation}
where the torsional spring torque $\tau_s$ is expressed as $k_s\theta_s$, with $k_s$ representing the spring stiffness coefficient. By substituting (\ref{eq:2.6}) and (\ref{eq:2.11}) into (\ref{eq:2.2}), $\tau_{in}$ is defined as: 
\begin{equation} 
    \tau_{in}=J_{coupler}(J_{strut}^{-1}\tau_s-W_w)
    \label{eq:2.12}
\end{equation}
Thus, $\Delta r$ and $\tau_{in}$ are determined as a function of $\theta_d$ in (\ref{eq:2.3}) and (\ref{eq:2.12}).

The presented mathematical relationships not only define the force- and geometry- based logic conditions that govern the MORPH wheel, but also formalize threshold-driven mechanical state switching, which is essential for realizing fully passive, sensor-less, and bidirectional radius adaptation. Unlike conventional CVT modeling focused solely on ratio prediction, this framework constitutes a mechanically executable decision structure, enabling the systematic tuning of geometric and compliant parameters to guarantee deterministic transformation behavior, high torque transmission capacity, and stable operation under varying load and locomotion conditions.

\begin{figure}[t]
    \centerline{\includegraphics[width=8cm]{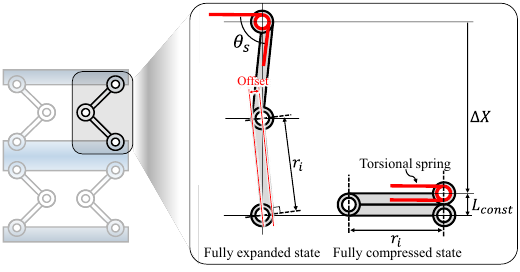}}
    \caption{Configuration of the connecting strut mechanism. Two symmetric slider–crank mechanisms are installed per tire segment, each equipped with a torsional spring (k = 2.14 N·mm/deg) at the upper joint.}
    \label{fig:fig_2.4}
\end{figure}

\section{Design of the MORPH Wheel}

\subsection{Design of Functional Components}

\begin{figure*}[t]
    \centerline{\includegraphics[width=16cm]{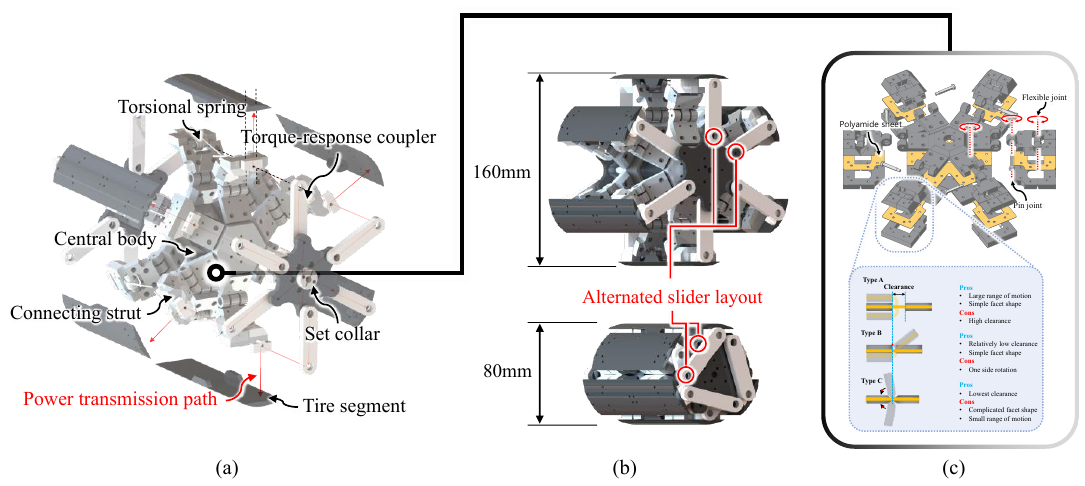}}
    \caption{Design and assembly of the MORPH wheel. (a) Disassembled components. (b) Assembled configuration with alternating slider layout. (c) Connecting strut connecting the central body and tire, featuring three joints. Flexible joint design types (A, B, C) are compared, with Type C fabricated via laser-cut polyamide sheets combined with 3D-printed links.}
    \label{fig:fig_3.1}
\end{figure*}

\begin{figure*}[t]
    \centerline{\includegraphics[width=15cm]{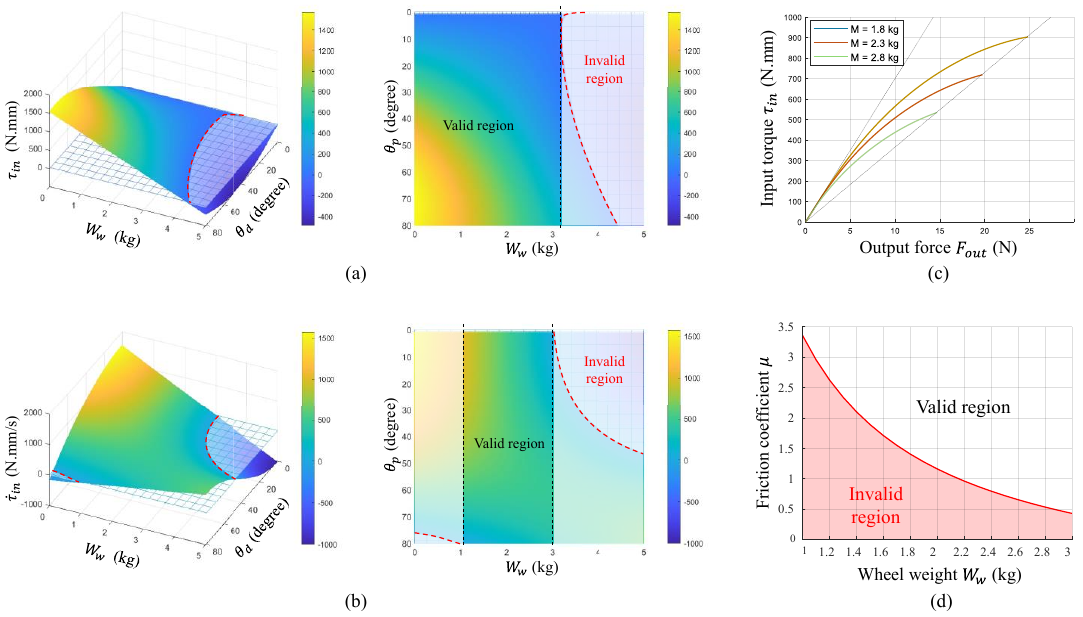}}
    \caption{Design process and constraints for wheel weight selection. (a) First design constraint: positive input torque $\tau_{in}$ across the drivetrain angle $\theta_d$, establishing upper bound $W_w \leq 3.1 \,\text{kg}$. (b) Second design constraint: positive torque derivative $\dot{\tau}_{in}$ ensuring contraction force generation, establishing lower bound $W_w \geq 1\,\text{kg}$ and refining upper bound to $W_w \leq 3 \,\text{kg}$. (c) Design trade-off between input torque and output force for candidate wheel weights (1.8 kg, 2.3 kg, 2.8 kg) within the feasible range. (d) Friction coefficient requirements for each design candidate to maintain traction without slip.}
    \label{fig:fig_3.2}
\end{figure*}

The MORPH wheel was designed to satisfy the following three primary objectives:\vspace{1mm}\\
\vspace{1mm}
(1) radius transformation ratio of at least 2.0 (80 $\rightarrow$ 40 mm),\\
\vspace{1mm}
(2) contraction force $\geq$ 10 N in the fully compressed state,\\
\vspace{1mm}
(3) symmetrical bidirectional motion.\\
The radius transformation ratio was set to 2.0, determined by examining actively controlled variable-radius wheels from prior research. Most existing active variable radius wheel exhibit transformation ratios of 2.0 or below \cite{20.lee2017origami, 21.lee2021high, 22.moger2024design}. To ensure that the MORPH wheel achieves performance comparable to or exceeding that of active counterparts, we adopted a transformation ratio of 2.0 as the design target. The contraction force capacity was also determined by referencing previous passive CVT wheels \cite{14.felton2014passive}. The existing passive CVT wheel design demonstrated a maximum contraction force capacity of 2 N. To validate the enhanced mechanical capabilities of the MORPH wheel, we established a significantly higher target contraction force of 10 N in the fully compressed state, representing a fivefold increase over the previous passive design. This substantial improvement in force capacity enables the MORPH wheel to support heavier payloads and operate effectively in more demanding terrain conditions.

In addition, the number of tire segments was also determined to ensure the vehicle's driving stability. Three key considerations were identified for achieving these design goals: the number of tire segments, the torque–response coupler, and the connecting strut. The number of tire segments significantly affects the stability of the robot platform. Unlike conventional tires with fixed geometry, the MORPH wheel requires segmentation to enable shape transformation. As shown in Fig.~\ref{fig:fig_2.3}(a), gaps arise between tire segments in the fully expanded state, causing variation in the central height $h_c$ of the wheel and thereby inducing a wheel displacement amplitude $A$. Since the number of tire segments $n$ directly affects this wheel displacement amplitude, selecting an appropriate value for $n$ is crucial. To define $A$, we established three assumptions: 1) the MORPH wheel forms a perfect circle with no gaps between tire segments in the fully compressed state; 2) the arc length of each tire segment is identical between the fully expanded and fully compressed states; 3) the arc of each tire segment in the fully expanded state forms a perfect circle with respect to the MORPH wheel's center. Based on these assumptions, the wheel displacement amplitude $A$ and normalized A are expressed as a function of $n$ and the maximum-to-minimum radius ratio $\rho_c$ as follows:
\begin{equation} 
\begin{aligned}
    A &=h_{c,max}-h_{c,min} \\
      &=h_{c,max}-r_{w,max} cos(\frac{\rho_c-1}{n\rho_c}\pi)
    \label{eq:2.13}
\end{aligned}
\end{equation}

\begin{equation} 
    \hat{A}=\frac{A}{h_{c,max}}*100
    \label{eq:2.14}
\end{equation}

Fig.~\ref{fig:fig_2.3}(b) illustrates vehicle normalized displacement amplitude ($\hat{A}$) calculated by \ref{eq:2.13} for $n = 3$–$10$ and $\rho_c = 1.2$–$2.0$ in increments of 0.2. The results show that vibration increases with $n$ and decreases with $\rho_c$. To maintain vehicle vibration below 5\% of the maximum wheel radius $r_{w,\max}$, the MORPH wheel was designed with six tire segments. Each segment arc was defined with respect to the wheel center in the fully compressed state, with an arc length of $41.8879\ \text{mm}$.

Achieving the target $\Delta r = 40$ mm requires proper design of the torque–response coupler. The coupler was implemented using multiple slider–crank mechanisms evenly distributed around the wheel (Fig.~\ref{fig:fig_2.2}(c)). The crank length $l_a$ and slider length $l_b$ were constrained by the following conditions:\vspace{1mm}\\
\vspace{1mm}
$\cdot$ singularity avoidance: $l_b$ $<$ 2$l_a$,\\
\vspace{1mm}
$\cdot$ required displacement: $l_b$ $\geq$ 40 mm,\\
\vspace{1mm}
$\cdot$ geometric feasibility: $l_a$+$l_b$ - clearance $<$ $r_{w,max}$\\
\vspace{0mm}
The maximum displacement of each slider–crank was designed to achieve the target $\Delta r$. The displacement range is $2l_a$ when $l_b > 2l_a$, and $l_b$ when $l_b < 2l_a$, where $l_a$ and $l_b$ denote crank and slider link lengths, respectively. Because $l_b > 2l_a$ can induce singularity, $l_b$ was constrained to $l_b < 2l_a$. Accordingly, $l_b$ was set to 40 mm to meet the 40 mm $\Delta r$ requirement. The total length $l_a + l_b$ must be shorter than $r_{w,\max}$ in the fully expanded state, with a 10 mm clearance required by physical constraints. Thus, $l_a$ was set to 30 mm. Under this configuration, achieving $\Delta r = 40$ mm requires a crank angle $\theta_p$ of 83.6°. However, due to interference among slider–crank units spaced at 60°, the maximum allowable angle is 45°. To resolve this limitation, sliders were alternately arranged in front and behind the cranks, increasing the allowable angle to 90°.

The connecting strut directly influences both the maximum displacement $\Delta r$ and spring deformation. For simplicity, each tire segment was equipped with two symmetric slider–crank mechanisms, each with a torsional spring mounted at the upper joint to impose wheel constraints (Fig.~\ref{fig:fig_2.4}). To achieve $\Delta r_{\max} = 40$ mm, the inner radius $r_i$ was determined such that the height difference between fully expanded and fully compressed states equals the sum of the design constraint length $L_{\text{const}}$ and $\Delta r_{\max}$. Here, $\theta_{s,\max} = 74^\circ$, $\theta_{s,\min} = 0^\circ$, and $L_{\text{const}} = 5$ mm were assumed. Because the symmetric slider–crank was offset to ensure $\theta_s = 0^\circ$ in the fully compressed state, $\theta_{s,\max}$ was less than 90°, thereby avoiding singularities. Substituting these values into Eq. x yielded $r_i \geq 24.6293$ mm; hence, $r_i$ was defined as 25 mm. Furthermore, The torsional spring stiffness was determined to be 2.14 N·mm/deg, based on a commercially available spring with a 90° angular configuration and a deformation capacity of approximately 90°, while considering physical constraints. A total of 12 torsional springs were installed in the MORPH wheel.



Figs.~\ref{fig:fig_3.1}(a) and (b) present the disassembled and assembled 3D CAD models of the MORPH wheel, respectively, designed based on the specified parameters. The main input shaft is directly coupled to the torque–response coupler via a set collar, while the central body is mounted on the shaft through bearings, ensuring that it remains unaffected by the main input. The MORPH wheel has a radius of 80 mm in the fully expanded state and 40 mm in the fully compressed state, with sliders arranged in an alternating layout, as shown in Fig. x(b).

Fig.~\ref{fig:fig_3.1}(c) depicts the link mechanism implementing the connecting strut. Each link mechanism connects the central body and tire segments, incorporating three joints. Torsional springs are installed at the connection points between links and tire segments to implement mechanical intelligence. Among these, the two joints connecting to the central body and wheel are configured as flexible joints. Flexible joints offer advantages in terms of manufacturing simplicity, maintenance ease, and cost-effectiveness, making them suitable for systems requiring numerous joints.

The flexible joint design is classified into three types as shown in Fig.~\ref{fig:fig_3.1}(c). Type A provides a large range of motion but has significant clearance, making it unsuitable for precise movements \cite{23.mintchev2019portable}, \cite{24.williams20224}. Type B, a variation of Type A, enables narrow clearance and precise movement but permits only unidirectional motion with limited stiffness. Type C offers the narrowest clearance and highest precision but features complex geometry and limited range of motion. In this study, Type C was selected to ensure high precision and stiffness, implemented by laser-cutting polyamide sheets and combining them with 3D-printed link components.

While flexible joints offer design advantages in manufacturing and maintenance simplicity, they are vulnerable to torsional force. To compensate for this limitation, the centrally positioned joint among the three joints was designed as a pin joint. Pin joints, manufactured with rigid bearings and pins, provide high precision and excellent torsional resistance. Therefore, by positioning the pin joint at the center where torsional force concentrates, the risk of flexible joint failure is prevented.


\subsection{Design of Wheel Weight}

The wheel weight represents a fundamental design parameter that directly influences the overall performance of the MORPH wheel. As shown in (\ref{eq:2.5}), the wheel weight $W_w$ defines the relationship between input torque and output force. In addition, the maximum static friction force governing wheel traction is determined by $W_w$ and the friction coefficient, and consequently, the wheel weight also determines the maximum contraction force. In this chapter, based on the mathematical model presented in Section II-C, the influence of $W_w$ on the performance of the MORPH wheel is analyzed, and the feasible range of wheel weight is identified.

Stable locomotion requires that $W_w$ satisfies two essential conditions simultaneously. The first condition concerns the input torque $\tau_{in}$, which remains positive over the entire $\theta_d$ domain. Negative $\tau_{in}$ values indicate that $W_w$ exceeds the spring resistance force $F_s$. In such intervals, torque is not applied for radius variation, and the MORPH wheel collapses into compression before reaching the next torque-demanding phase. This unintended radius variation reduces the attainable transmission ratio range and restricts the performance of the variable transmission. The input torque $\tau_{in}$ as a function of $W_w$ and $\theta_d$ is calculated from (\ref{eq:2.12}) and presented in Fig.~\ref{fig:fig_3.2}. When $\tau_{in}$ remains positive throughout the entire $\theta_d$ domain, the mechanical behavior logic operates properly. Under this condition, $W_w \leq 3.1$ kg defines a valid region, and values above this threshold are considered invalid.

The second condition concerns the derivative of input torque with respect to $\theta_d$, denoted $\dot{\tau}_{in}$. This derivative needs to remain positive across the full $\theta_d$ domain. Negative values of $\dot{\tau}{in}$ indicate a decreasing torque requirement for radius variation, which reduces the generation of contraction force. This effect, as in the first condition, narrows the achievable TR range and restricts the effectiveness of the variable transmission. Numerical differentiation using MATLAB was employed to calculate $\dot{\tau}_{in}$, since its highly nonlinear form limits analytical treatment. As shown in Fig. \ref{fig:fig_3.2}(b), negative $\dot{\tau}_{in}$ values appear for $W_w < 1$ kg and $W_w > 3$ kg, indicating invalid regions. Accordingly, the valid region for proper mechanical behavior logic is defined as $1 \leq W_w \leq 3$ kg. By considering both conditions simultaneously, $W_w$ remains effective only when located within this range.

Within the identified range, three representative wheel weights of 1.8 kg, 2.3 kg, and 2.8 kg are substituted into (\ref{eq:2.11}) to evaluate the relationship between input torque and output force, as shown in Fig.~\ref{fig:fig_3.2}(c). For $W_w = 1.8$ kg, the input torque reaches approximately 900 N·mm at the fully compressed state, producing an output force of about 21.74 N. For $W_w = 2.3$ kg and $W_w = 2.8$ kg, the fully compressed state occurs at input torques of 710 N·mm and 530 N·mm, respectively, with corresponding output forces of 17.02 N and 12.24 N.

The maximum static friction force depends on both $W_w$ and the friction coefficient $\mu$. To prevent slip, the feasible region of $\mu$ is defined as a function of $W_w$. Considering wheel dynamics, slip does not occur when $F_{out} \leq \mu W_w$, which indicates that stable locomotion requires $\mu \geq F_{out}/W_w$. Substituting (\ref{eq:2.11}) into this condition produces the valid region, illustrated as the white area in Fig.~\ref{fig:fig_3.2}(d), where the wheel maintains traction without slip.

Beyond the effect of wheel weight, the torsional spring stiffness $k_s$ also plays a critical role in determining the output force. Fig. x illustrates how the maximum output force $F_{out,max}$ varies with the torsional stiffness $k_s$ for a fixed geometry and wheel weight (2.8 kg). As $k_s$ increases, the required input torque $\tau_{in,max}$ monotonically, extending the achievable transmission ratio range but also increasing the torque demand on the actuator. This trade-off directly links an easily tunable design parameter ($k_s$) to a key performance metric ($F_{out,max}$), and guides the selection of spring stiffness for a given actuator capability.

\begin{figure}[t]
    \centerline{\includegraphics[width=7.13cm]{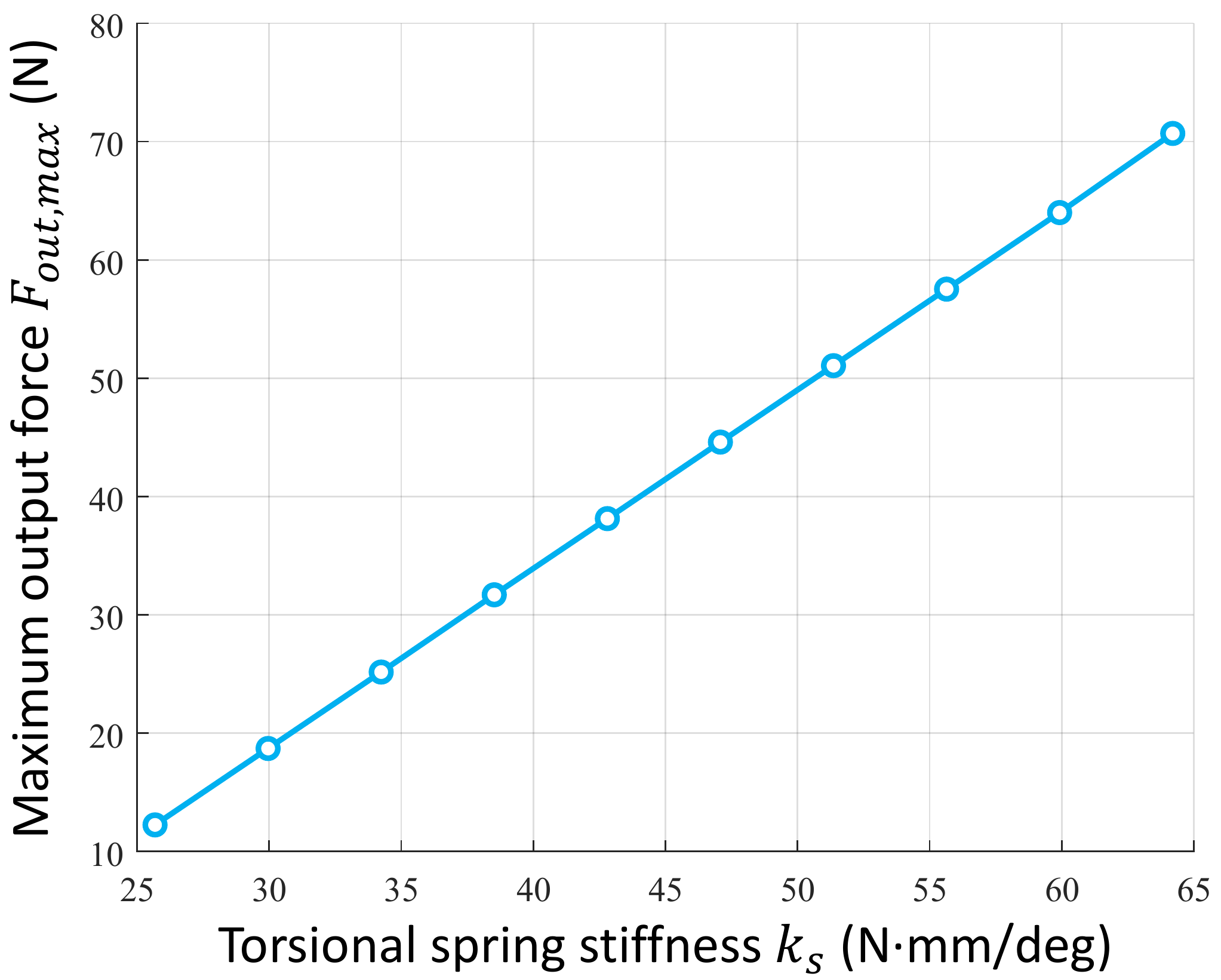}}
    \caption{Effect of torsional spring stiffness $k_s$ on maximum output force $F_{out,max}$.}
    \label{fig:fig_3.3}
\end{figure}

\begin{figure}[t]
    \centerline{\includegraphics[width=6cm]{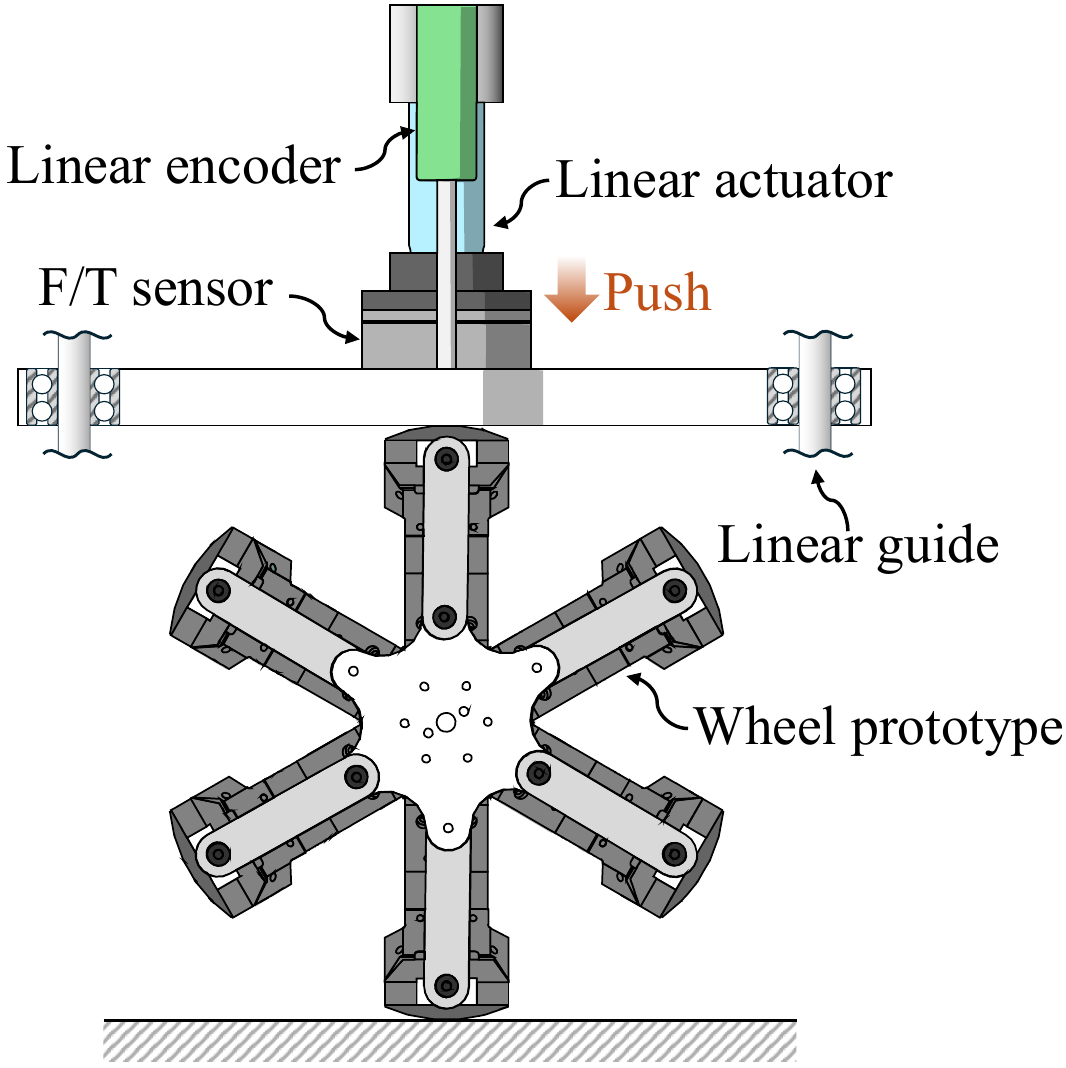}}
    \caption{Experimental setup for spring resistance force validation.}
    \label{fig:fig_4.1}
\end{figure}

\begin{figure}[t]
    \centerline{\includegraphics[width=7cm]{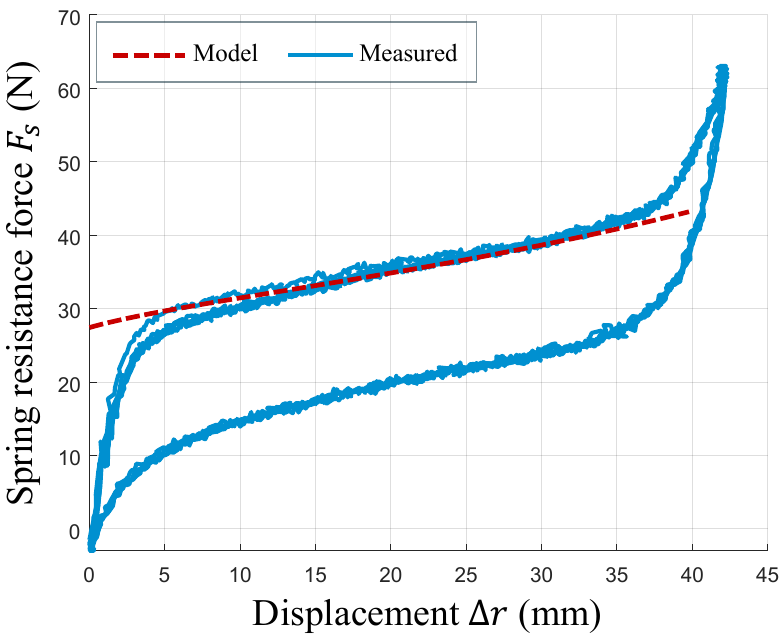}}
    \caption{Spring resistance force validation results. Measured force ($F_s$) versus radius variation ($\Delta r$) represented as mean and standard deviation across five trials, compared with model-predicted force ($F_c$).}
    \label{fig:fig_4.2}
\end{figure}

\section{Evaluation of Mechanical Behavior Logic Operating Conditions}
This section presents quantitative experimental results to evaluate the performance of the MORPH wheel. The mechanism aims to implement continuous transmission functionality by passively adjusting wheel radius according to input torque magnitude without requiring active control. To validate this concept, radius variations were measured under different compression forces and input torque conditions, and the results were compared with theoretical models.

\begin{figure}[t]
    \centerline{\includegraphics[width=9cm]{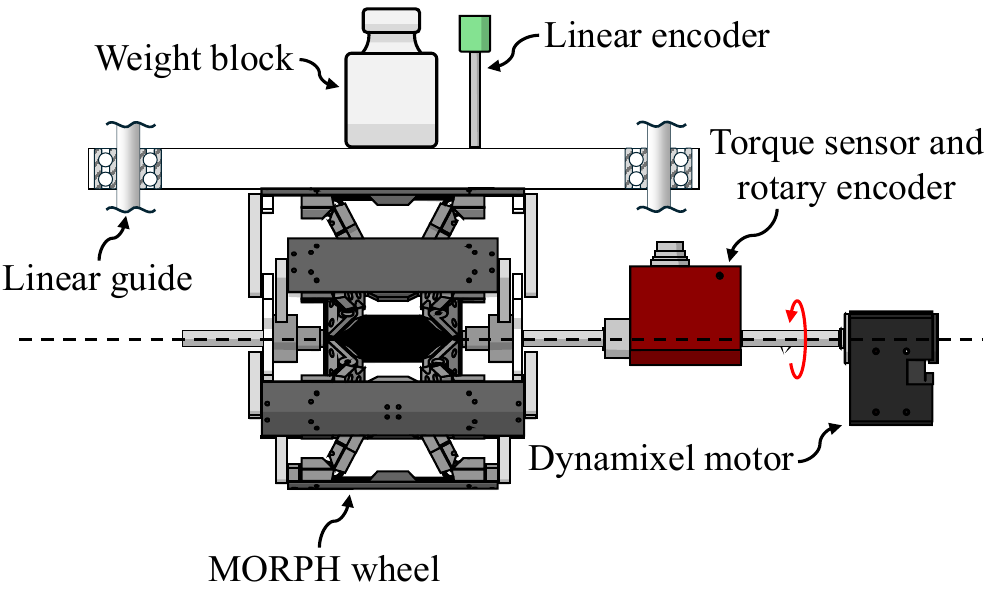}}
    \caption{Experimental setup for mechanical behavior logic validation.}
    \label{fig:fig_4.3}
\end{figure}

\begin{figure*}[t]
    \centerline{\includegraphics[width=16cm]{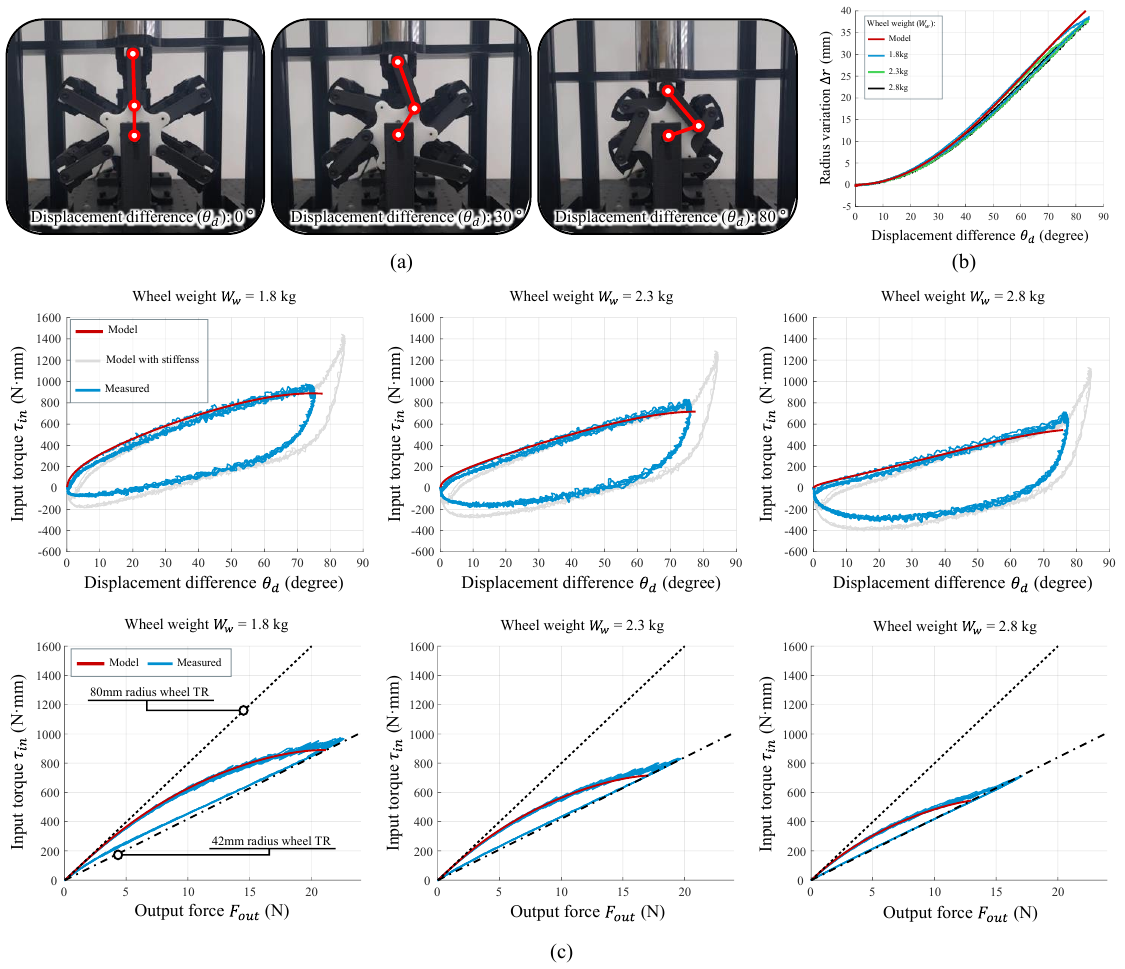}}
    \caption{Experimental validation of the MORPH wheel kinematics and kinetics. (a) Wheel prototype transformation during testing showing drivetrain angle $\theta_d$ variation from 0° to 84°. (b) Measured wheel kinematics showing radius variation $\Delta r$ versus $\theta_d$ for three wheel weights (1.8, 2.3, 2.8 kg), represented as mean with standard deviation from five trials. (c) Wheel kinetics showing input torque $\tau_{in}$ versus $\Delta r$, where red dashed line represents theoretical model, gray line indicates recalculated values using measured spring resistance force ($F_s$), and blue line shows measured results.}
    \label{fig:fig_4.4}
\end{figure*}

\subsection{Experimental Evaluation of Spring Resistance Force}

The torsional spring is the primary element defining the MORPH wheel’s mechanical behavior logic by shaping the resistance force profile applied to the mechanism. Its accuracy directly governs the wheel’s performance as a variable transmission, since deviations in stiffness alter the programmed transmission-ratio trajectory and the $\tau_{in}$–$F_{out}$ mapping. This section presents a quantitative verification that the installed spring reproduces the designed resistance force $F_s$ across the specified deflection range.

The experimental setup consisted of the MORPH wheel prototype, a linear guide, a force/torque sensor (Mini45, ATI Industrial Automation), a linear encoder (PZ34-A-150, GEFRAN), and a linear actuator (LM4075, Motorbank) as shown in Fig.~\ref{fig:fig_4.1}. To constrain rotation, one tire segment of the MORPH wheel was mechanically fixed against the linear guide. The force sensor was mounted on the guide, and the linear actuator was positioned above it to apply compression to the wheel. The resulting spring resistance force $F_s$ was measured by the force sensor, while the corresponding radius variation $\Delta r$ was captured by the linear encoder. During testing, $\Delta r$ was gradually increased from 0 to 42 mm, and $F_s$ was sampled at 100 Hz. To minimize disturbances and measurement noise, each experiment was repeated five times, ensuring statistical reliability for quantitative analysis.

Across all five trials, the measured spring resistance force closely matched the model-predicted spring resistance force $F_s$ derived from (\ref{eq:2.11}), showing strong qualitative agreement (Fig.~\ref{fig:fig_4.2}). However, within the initial 0–5 mm range of radius variation, the measured compression force deviated from the theoretical model by approximately 16–27 N. This discrepancy is attributed to clearance in the joints of the slider–crank mechanism. As the radius begins to contract, the joint clearance is gradually reduced, which delays the buildup of resistance force. This phenomenon is further discussed in the following section.

At larger deformations, a sharp exponential increase in measured force was observed beyond $\Delta r = 38$ mm. This increase results from mechanical constraints of the structure, marking the range in which further deformation of the MORPH wheel is physically restricted. Despite a deviation of approximately 2 mm from the design specification, the spring resistance force was verified to accurately realize the intended mechanical behavior logic condition throughout the effective operating range of 5–38 mm.

Finally, hysteresis was observed between the compression and expansion phases. In the torsional spring, clearance exists at the mounting interface of the spring’s rotational axis. This clearance may cause the effective rotation center to shift between compression and expansion, producing the observed hysteresis. Such effects are expected to be mitigated through higher-precision fabrication in future iterations of the MORPH wheel.

\subsection{Experimental Evaluation of Mechanical Behavior Logic}

A primary contribution of the MORPH wheel is the precisely and accurately programmed TR variation embedded in its mechanical design. This section presents a quantitative experimental evaluation of the wheel’s performance by characterizing (i) the radius variation $\Delta r$ as a function of displacement difference $\theta_{d}$, and (ii) the transmission mapping between $\tau_{in}$ and the resulting output force $F_{out}$. The objectives are to confirm that the realized $\tau_{in}(\theta_d)$ trajectory adheres to the designed TR profile and to verify that the $\tau_{in}$–$F_{out}$ relationship demonstrates stable operation across the intended variable transmission range.

The experimental setup consisted of the MORPH wheel prototype, a linear guide, a torque sensor (TRS605, FUTEK), a rotary encoder (360 pulses/rev), a linear encoder (PZ34-A-150, GEFRAN), and a Dynamixel motor (XM540-W270-R, Robotis) as indicated in Fig.~\ref{fig:fig_4.3}. As in the spring resistance force experiments, one tire segment was mechanically fixed against the linear guide to prevent wheel rotation. A weight block was placed on the linear guide to represent the effective wheel weight $W_w$. Input torque was generated using the Dynamixel motor, which was coupled to the MORPH wheel’s torque-response coupler. The torque sensor and rotary encoder positioned between the motor and the wheel measured the applied input torque $\tau_{in}$ and the displacement difference $\theta_d$, respectively, while the linear encoder captured the corresponding radius variation $\Delta r$.
During testing, $\theta_d$ was gradually increased from 0° to 84° as illustrated in Fig.~\ref{fig:fig_4.4}(a), and $\tau_{in}$, $\theta_d$, and $\Delta r$ were sampled at 100 Hz. To examine the effect of different wheel weights and to compare against theoretical predictions, experiments were conducted with $W_w$ set to 1.8, 2.3, and 2.8 kg (including the mass of the linear guide and sensors). Each condition was repeated five times to ensure statistical validity and to minimize disturbances or measurement noise.

Fig.~\ref{fig:fig_4.4}(b) presents the measured kinematics of the MORPH wheel based on the $\theta_d$–$\Delta r$ data. The experimental results across all weight conditions qualitatively matched the model predictions. However, the deviation from theoretical values increased with larger wheel weights. This discrepancy is attributed to deformation of the plastic crank and slider components under higher loads.
Fig.~\ref{fig:fig_4.4}(c) further illustrates the wheel kinetics derived from $\tau_{in}$–$\Delta r$ measurements. The red solid line represents theoretical values from the model, the gray solid line indicates recalculated values using the measured spring resistance force $F_s$, and the blue solid line shows directly measured results. Across all trials, strong agreement was observed between the theoratical model and measured data. Notably, the error in the 0–5 mm range was significantly smaller than that observed in the spring resistance force experiments. This improvement is likely due to preload effects from the wheel weight, which eliminated joint clearance. Consequently, errors in $F_s$ within the 0–5 mm region can be considered negligible when the MORPH wheel is applied to a robot platform.

Across all conditions, hysteresis was consistently observed between the compression and expansion phases. The measured hysteresis closely matched the behavior of the model incorporating the experimentally obtained stiffness, indicating that the effect arises from clearance at the interface between the spring and the rotational axis, as also noted in the spring resistance experiments. In summary, the results confirm that the MORPH wheel reliably realizes the intended variable transmission behavior within the effective operating range of $\Delta r = 0$–38 mm, validating the accuracy of the mechanically programmed TR variation.

\begin{figure}[t]
    \centerline{\includegraphics[width=9cm]{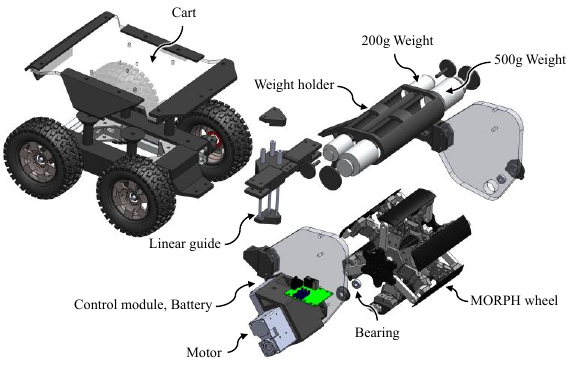}}
    \caption{Robot platform equipped with the MORPH wheel.}
    \label{fig:fig_5.0}
\end{figure}


\begin{figure*}[t]
    \centerline{\includegraphics[width=16cm]{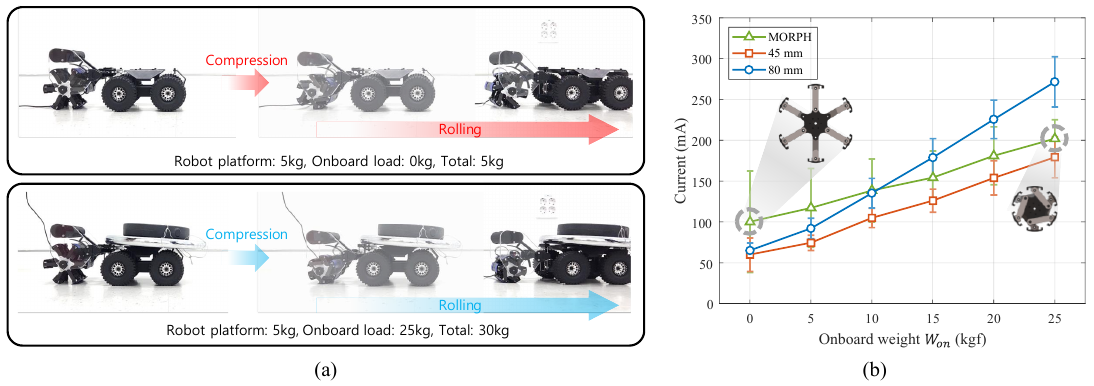}}
    \caption{Driving performance evaluation under varying onboard loads. (a) Robot operation showing wheel radius variation with 0 kg and 25 kg onboard load ($W_{on}$). (b) Motor driving current versus onboard load (mean and standard deviation) for variable radius wheel and fixed radius wheels (80 mm and 45 mm radius) at 60 rpm.}
    \label{fig:fig_5.1}
\end{figure*}

\begin{figure}[t]
    \centerline{\includegraphics[width=8cm]{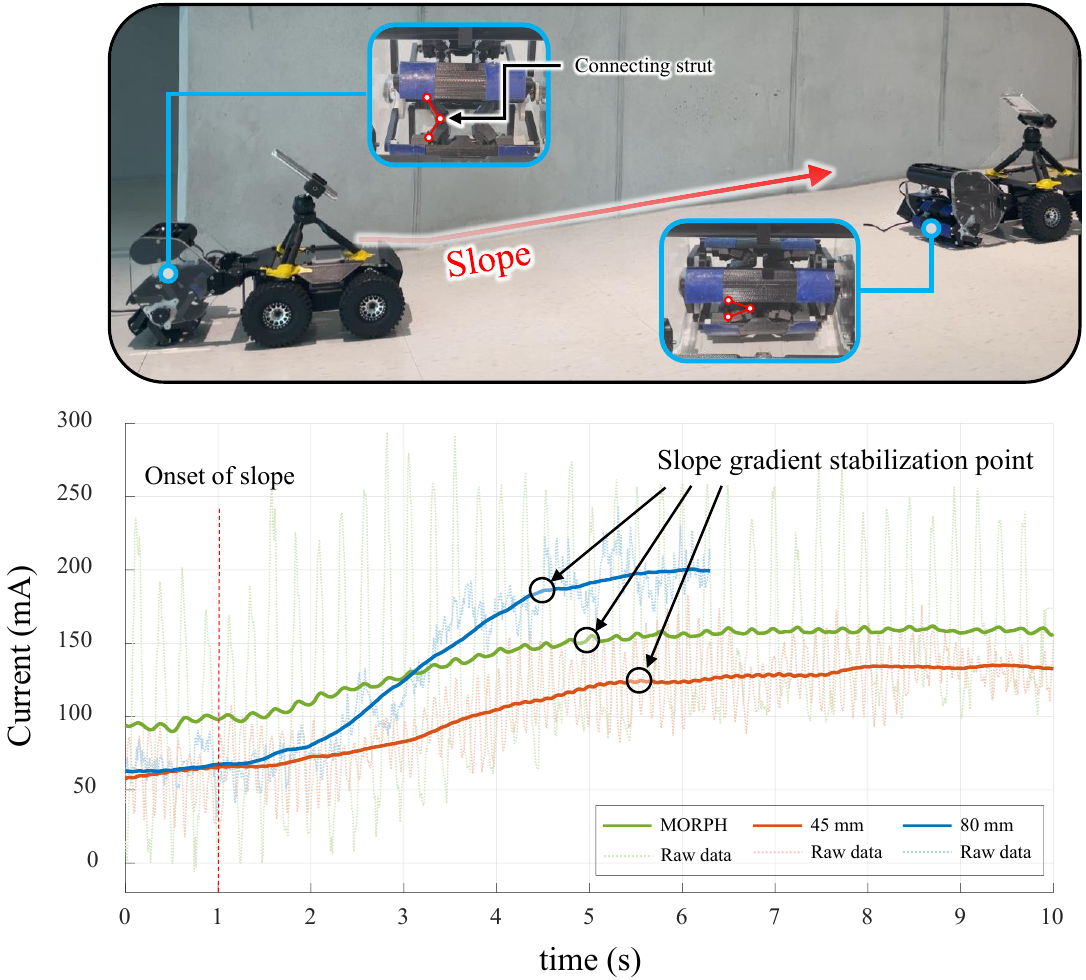}}
    \caption{Comparative performance evaluation on continuous flat-to-inclined terrain. Motor driving current and locomotion speed for three wheel configurations: fixed maximum-radius wheel (80 mm, Blue solid line), fixed minimum-radius wheel (45 mm, Red solid line), and the MORPH wheel (Green solid line).}
    \label{fig:fig_5.2}
\end{figure}

\section{Robot Platform and Demonstration}

\begin{table}[h]
\caption{Specifications of Passive Variable Radius Wheel and Robot Platform}
\label{tab:wheel_spec}
\centering
\begin{tabular}{ll}
\hline
\multicolumn{2}{c}{\textbf{MORPH Wheel}} \\
\hline
Maximum radius      & 80\,mm \\
Minimum radius      & 42\,mm \\
Transmission ratio  & 160\% \\
Material            & Acrylonitrile Butadiene Styrene (ABS) \\
Flexible joint      & Polyamide sheet (0.1\,mm) \\
Spring stiffness    & 25.68 (2.14 * 12)\,N$\cdot$mm/deg \\
\hline
\multicolumn{2}{c}{\textbf{Robot Platform}} \\
\hline
Motor               & Dynamixel XM540-W270 \\
Wheel weight        & 1.1\,kg + 1.4\,kg (Additional weight) \\
Cart weight         & 2.5\,kg \\
Total weight        & 5\,kg \\
\hline
\end{tabular}
\end{table}

\begin{figure*}[t]
    \centerline{\includegraphics[width=17cm]{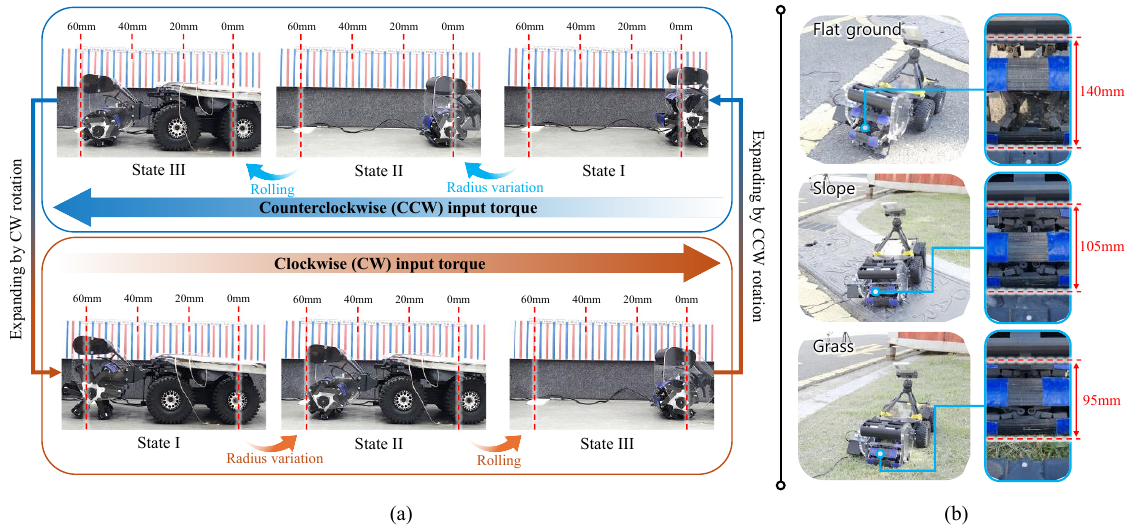}}
    \caption{(a) Bidirectional operation demonstration of the MORPH wheel. Sequential radius transformation during counterclockwise (CCW, red arrow) and clockwise (CW, blue arrow) rotation. (b) Natural terrain adaptability demonstration of the MORPH wheel. Robot platform operation across three distinct terrain conditions: flat ground, sloped terrain, and grass surface.}
    \label{fig:fig_5.3}
\end{figure*}

To validate the practical applicability of the MORPH wheel, a dedicated robotic platform was developed and a series of demonstrations were conducted. The platform was designed to directly reveal the wheel’s characteristic behavior—namely, its passive variation in radius in response to applied torque under actual driving conditions. It was equipped with onboard electronics and batteries, allowing fully independent operation without reliance on external power sources. The objective of this section is to verify the MORPH wheel’s mechanically programmed functionality at the system level by examining how its transformation translates into actual robotic locomotion performance.

For experimental verification in realistic driving environments, the platform was constructed with two primary modules: a wheel assembly incorporating the MORPH wheels and a cart for carrying loads (Fig.~\ref{fig:fig_5.0}). The two modules were connected via linear guides, ensuring smooth accommodation of wheel-radius changes during operation. The wheel assembly integrated drive motors for locomotion, a microcontroller for control tasks, and a battery for power supply, while the cart was designed to hold additional weight blocks that imposed varying load conditions. This configuration enabled systematic evaluation of both the wheel’s radius transformation and the resulting locomotion performance under diverse operating scenarios. Furthermore, by monitoring the motor driving current, the system provided a quantitative basis for analyzing performance throughout the demonstrations.


\subsection{Evaluation of Radius Variation Characteristics Under Onboard Weight Conditions}

To assess the performance of the MORPH wheel, a robot platform was tested with three wheel types: a fixed maximum-radius wheel (80mm), a fixed minimum-radius wheel (45mm), and the MORPH wheel. Driving experiments were conducted for each wheel type with the onboard load $W_{on}$ incrementally increased in 5 kg intervals. The wheel rotation speed was maintained at 60 rpm across all experimental conditions, with each condition tested for 6 seconds while measuring motor driving current. For performance comparison, static wheels fixed at maximum radius (80 mm) and minimum radius (45 mm) were evaluated under identical conditions.

Fig.~\ref{fig:fig_5.1}(a) illustrates the robot operation under no additional load and with 25 kg onboard load $W_{on}$. The experimental results confirmed that the wheel radius distinctly decreased as the load increased. As shown in Fig.~\ref{fig:fig_5.1}(b), the MORPH wheel exhibited relatively smaller current increases compared to static wheels, indicating that radius variation effectively reduced the required input torque ($\tau_{in}$) under identical onboard load $W_{on}$ conditions. However, when onboard load $W_{on}$ were 5 kg and 10 kg, the MORPH wheel consumed more current than the maximum radius wheel. This phenomenon was attributed to driving vibrations caused by the wheel surface geometry comprising curved and flat surfaces, resulting in larger standard deviations in current consumption compared to static wheels. Nevertheless, as the load increased, the wheel geometry approached a circular shape, reducing vibrations and consequently decreasing the standard deviation.

\subsection{Evaluation of Radius Variation Characteristics Under Inclined Slope Conditions}

Under the same experimental conditions, inclined slope tests were performed with the two fixed-radius wheels (80mm and 45mm) and the MORPH wheel. The test environment consisted of a continuous terrain composed of flat ground followed by an inclined slope. While fixed-radius wheels offer advantages under specific conditions, they lack the adaptability required to accommodate varying environments. Accordingly, the two fixed wheels served as baselines against which the MORPH wheel’s adaptive performance was evaluated.

As illustrated in Fig.~\ref{fig:fig_5.2}, the maximum-radius wheel exhibited a sharp increase in motor current, rising from approximately 50 mA on flat ground to nearly 200 mA upon entering the slope. In contrast, the MORPH wheel maintained narrow current variation range during the transition from ground to inclined terrain. This result indicates that, as with existing passive CVTs \cite{7.belter2014passively, 9.kim2020elastomeric}, the MORPH wheel allows a properly selected actuator to operate within its peak efficiency range across varying operating conditions, since its torque-responsive geometry continuously adjusts the effective radius to deliver an output-matched transmission ratio while minimizing variation in actuator input. Consequently, the MORPH wheel demonstrates the potential to improve overall locomotion efficiency without reliance on additional actuators, sensors, and active control.

Locomotion speed was also examined as a key performance metric. When traversing the inclined section, the robot equipped with the maximum-radius wheel required approximately 3.5 s to reach the stabilized portion of the slope, whereas the minimum-radius wheel required 4.5 s. The MORPH wheel achieved an intermediate value of about 4.0 s. These results highlight that, unlike the minimum-radius wheel which simply reduces current consumption, the MORPH wheel balances the advantages of both extremes by maintaining energy efficiency while preserving competitive locomotion performance under changing terrain conditions.

In summary, the MORPH wheel effectively suppressed excessive current consumption and maintained motor operation within the narrow efficiency range across diverse terrains. These findings experimentally verify that the MORPH wheel secures both energy efficiency and performance enhancement when compared with conventional fixed-radius wheels.



\subsection{Evaluation of Bidirectional Capability}

A key feature of the MORPH wheel is its bidirectional capability, which ensures reliable operation under both counterclockwise (CCW) and clockwise (CW) input conditions. The origami pattern in previous passive CVT for wheeled robots \cite{14.felton2014passive} is inherently unidirectional because its geometry is preprogrammed to fold only under forward torque. When torque is applied in this direction, the faces wrap inward and reduce the radius, but the crease configuration does not permit the outward folding required for reverse-torque deformation. As a result, the mechanism can modulate radius in only one direction, making bidirectional transmission operation impossible.

In contrast to preprogrammed pattern-based approaches, the MORPH wheel utilizes a slider-crank mechanism as a torque-responsive coupler. The inherent kinematic characteristics of the slider-crank mechanism enable it to generate contraction motion regardless of the crank's rotational direction around the singularity point. This mechanical property ensures that the MORPH wheel consistently transforms toward a reduced radius configuration in response to applied torque, thereby achieving bidirectional transformation capability.

The objective of this evaluation is to demonstrate that the wheel consistently performs variable-radius transformations in both rotational directions and to verify that the radius variation process is reproduced identically when the driving direction is reversed. This validation confirms that the MORPH wheel can stably realize its intended variable transmission function regardless of input orientation. This section presents the experimental procedure used to examine the bidirectional behavior and the corresponding results.

The demonstration was conducted on a single MORPH wheel under two distinct input conditions: CCW and CW rotation. 
In the CCW rotation condition, two distinct phases were observed as illustrated in the blue box and blue arrow in Fig.~\ref{fig:fig_5.3}(a). Under the condition $F_{out} \leq F_{res}$,the input torque triggers the transformation of the MORPH wheel from state I to state II. Once this threshold was reached, and $F_{out} > F_{res}$, the input torque was transmitted to wheel rotation without further radius variation, driving the robot forward by about 47.5 mm (from state II to state III) without further radius variation.

Subsequently, the driving direction was reversed to CW, returning the MORPH wheel from fully compressed state to the fully expanded state. The same procedure was then repeated in the CW direction, as shown by the red arrow in Fig.~\ref{fig:fig_5.3}(a), allowing the wheel to retrace the prior process and return to its initial position. In both cases, the wheel started from the fully expanded state, passed through the fully compressed state, and completed the rolling sequence, enabling a direct comparison of the radius variation process between the two rotational directions.

In conclusion, the MORPH wheel exhibited an identical radius variation process under both CCW and CW conditions. These results confirm that the structural symmetry of the MORPH wheel is faithfully manifested in physical operation, such that the radius transformation is consistently reproduced irrespective of input direction. Consequently, the MORPH wheel can be visually verified to maintain stable variable transmission performance in bidirectional driving scenarios, including both forward and reverse locomotion.

\subsection{Evaluation of Driving Performance on Natural Terrain Environments}



To assess the mechanical adaptability of the MORPH wheel on unstructured ground, experiments were conducted in natural terrain environments. Experiments were conducted on irregular slopes with mixed gravel and grass conditions. Fig.~\ref{fig:fig_5.2}(b) visually demonstrated the mechanism's adaptability through continuous wheel radius deformation according to terrain conditions. 
The experimental results showed that the MORPH mechanism autonomously adapted to external condition changes without sensors or active control, maintaining driving stability.
This demonstrates that the MORPH wheel can be effectively utilized as a passive CVT for robot mobility platforms with practicality and reliability in various complex environments.

\section{Conclusion and Future Work}

In this paper, we presented the MORPH wheel, a fully passive, torque-responsive, and mechanically programmed variable-radius wheel designed to address the unmet requirements of wheeled robotic locomotion. Unlike prior passive CVT approaches that were limited by unidirectional operation, low torque capacity, or shape-dependent deformation, the MORPH wheel enables continuous bidirectional operation, high and repeatable torque transmission, and deterministic transformation behavior without reliance on sensor, actuator,s or electronic control.

We established a comprehensive mechanical behavior logic model that defines threshold-based mode switching, and validated its accuracy through bench-top mechanical evaluation and mobile robot demonstrations across varied load and terrain conditions. The results confirm that mechanical programming serves as a viable design paradigm for embedding context-dependent, passive adaptation
directly into physical hardware without increasing system complexity, energy consumption, or vulnerability to environmental conditions.

Moving forward, we will refine the geometric continuity of the tire segments to achieve improved rolling dynamics, explore material and fabrication strategies for long-term fatigue resistance, and integrate the MORPH wheel into multi-wheel robotic platforms to evaluate real-world operational benefits.We believe that the concept of mechanically programmed adaptation introduced in this work can extend beyond variable-radius wheels to enable the next generation of electronics-free intelligent mechanisms for resilience-critical robotics.

\bibliographystyle{IEEEtran}
\bibliography{ref.bib}

\vfill

\end{document}